\documentclass[pdflatex,sn-mathphys-num]{sn-jnl}

\usepackage{graphicx}%
\usepackage{multirow}%
\usepackage{amsmath,amssymb,amsfonts}%
\usepackage{amsthm}%
\usepackage{mathrsfs}%
\usepackage[title]{appendix}%
\usepackage{xcolor}%
\usepackage{textcomp}%
\usepackage{manyfoot}%
\usepackage{booktabs}%
\usepackage{algorithm}%
\usepackage{algorithmicx}%
\usepackage{algpseudocode}%
\usepackage{listings}%
\usepackage{subcaption}%
\usepackage[section]{placeins} 

\theoremstyle{thmstyleone}%
\theoremstyle{thmstyletwo}%

\theoremstyle{thmstylethree}%

\begin{document}

\title[Visualizing the Invisible: Enhancing Radiologist Performance in Breast
Mammography via Task-Driven Chromatic Encoding]{Visualizing the Invisible: Enhancing Radiologist Performance in Breast Mammography via Task-Driven Chromatic Encoding}

\author[1]{\fnm{Hui} \sur{Ye}}\email{hui.ye@siat.ac.cn}
\equalcont{These authors contributed equally to this work.}

\author[1,2]{\fnm{Shilong} \sur{Yang}}\email{sl.yang0518@gmail.com}
\equalcont{These authors contributed equally to this work.}

\author*[1]{\fnm{Chulong} \sur{Zhang}}\email{chulong.zhang@duke.edu}

\author[2,3]{\fnm{Yexuan} \sur{Xing}}\email{xyx240311@163.com}

\author[3]{\fnm{Juan} \sur{Yu}}\email{yujuan0072@qq.com}

\author*[4]{\fnm{Yaoqin} \sur{Xie}}\email{xieyaoqin@suat-sz.edu.cn}

\author*[5]{\fnm{Wei} \sur{Zhang}}\email{holly2yang@126.com}

\affil[1]{\orgname{Shenzhen Institutes of Advanced Technology, Chinese Academy of Sciences},
\orgaddress{\street{1068 Xueyuan Avenue, Shenzhen University Town}, \city{Shenzhen}, \postcode{518055}, \state{Guangdong}, \country{China}}}

\affil[2]{\orgname{Shenzhen University},
\orgaddress{\street{No. 3688 Nanhai Avenue, Nanshan District}, \city{Shenzhen}, \postcode{518060}, \state{Guangdong}, \country{China}}}

\affil[3]{\orgdiv{Department of Radiology},
\orgname{The First Affiliated Hospital of Shenzhen University, Health Science Center, Shenzhen Second People's Hospital},
\orgaddress{\street{3002 SunGangXi Road}, \city{Shenzhen}, \postcode{518035}, \state{Guangdong}, \country{China}}}

\affil[4]{\orgname{Shenzhen University of Advanced Technology},
\orgaddress{\street{1 Gongchang Road}, \city{Shenzhen}, \state{Guangdong}, \country{China}}}

\affil[5]{\orgdiv{Department of Radiology},
\orgname{Liuzhou People's Hospital, Guangxi Medical University},
\orgaddress{\street{No. 8 Wenchang Road}, \city{Liuzhou}, \postcode{545006}, \state{Guangxi}, \country{China}}}



\abstract{
\textbf{Purpose:}
Mammography screening is less sensitive in dense breasts, where tissue overlap and subtle findings increase perceptual difficulty and inter-observer variability. We present \textbf{MammoColor}, an end-to-end framework incorporating a \textbf{Task-Driven Chromatic Encoding (TDCE)} module to convert single-channel mammograms into task-informed, \emph{TDCE-encoded} views for reader-oriented visual augmentation.

\textbf{Materials and Methods:}
MammoColor integrates a lightweight TDCE module with a BI-RADS triage classifier and was trained end-to-end on VinDr-Mammo to learn a task-optimized chromatic mapping that improves benign--malignant separability. Model performance was evaluated on an internal test set, two public datasets (CBIS-DDSM and INBreast), and three independent external clinical cohorts (multicenter validation). We also conducted a multi-reader, multi-case (MRMC) observer study with a washout period. Radiologists (junior, intermediate, senior) interpreted cases under three conditions: (1) grayscale-only, (2) TDCE-only (TDCE-encoded images only), and (3) side-by-side grayscale + TDCE. Outcomes included AUC, sensitivity, specificity, and inter-reader agreement ($\kappa$).

\textbf{Results:}
On VinDr-Mammo, MammoColor improved AUC from 0.7669 to 0.8461 ($P=0.004$). Performance showed adaptability across public datasets and external multicenter cohorts, supporting robustness under domain shift, particularly in FFDM cohorts. In prespecified subgroup analyses, the gain was larger in dense breasts (AUC 0.749 to 0.835). In the MRMC study, TDCE-encoded images improved specificity (0.90 to 0.96; P=0.052) while maintaining comparable sensitivity, suggesting potential utility in reducing false-positive recalls in screening triage..

\textbf{Conclusion:}
MammoColor, via TDCE, provides a task-optimized chromatic representation that bridges the grayscale-to-RGB gap as an intuitive, human-centric augmentation. TDCE-encoded visualization effectively promotes the accuracy and efficiency of mammography triage through improved perceptual salience, offering a practical tool for visual augmentation in routine clinical practice.
}

\maketitle

\section{Introduction}

Breast cancer remains the most commonly diagnosed malignancy and a leading cause of cancer-related mortality among women worldwide \cite{Bray2024}. Early detection through population-based screening is therefore central to reducing disease burden, and full-field digital mammography (FFDM) remains the standard modality. However, mammographic interpretation is inherently a demanding perceptual task. In dense breasts, superimposed fibroglandular tissue produces a pronounced masking effect that can conceal malignancies and increase false-negative interpretations \cite{Kolb2002}. Moreover, subtle findings---such as microcalcification clusters or early architectural distortion---may be difficult to distinguish from complex background parenchymal patterns, contributing to substantial inter-observer variability \cite{Elmore1994}. This variability is clinically consequential and strongly associated with reader experience \cite{Elmore1994}, motivating tools that reduce perceptual ambiguity and support more standardized decision-making.

Artificial intelligence (AI) systems have advanced rapidly as aids for mammography interpretation. Earlier computer-aided detection/diagnosis systems (CAD/CADx) provided limited clinical benefit due to high false-positive rates and poor interpretability \cite{Fenton2007}. Modern deep learning approaches show broad promise for medical image analysis \cite{topol2019}, yet two bottlenecks remain prominent in mammography. First, a \emph{representation mismatch} arises because widely used backbones are typically pretrained on three-channel natural images (e.g., ImageNet), whereas mammograms are single-channel grayscale \cite{Deng2009ImageNet,Xie2019}. Simple channel replication enables architectural compatibility but fails to exploit the full potential of  transferable RGB-pretrained features. Second, and more critical for clinical translation, many AI systems remain interaction-limited: they often output a ``black-box'' probability score or sparse localization cues, which may not directly improve radiologists' perception of subtle, low-contrast findings embedded in dense anatomy.

\emph{Chromatic encoding} offers an alternative strategy by transforming grayscale images into a multi-channel representation that can better align with RGB-pretrained models and, potentially, with human visual perception. Prior medical-imaging approaches are often handcrafted or intensity-driven (i.e., fixed pseudo-color colormaps) \cite{Zeng2020,Li2020}. Because such mappings are image-agnostic and not optimized for diagnostic objectives, they may amplify irrelevant structures or introduce inconsistent visual cues. In contrast, learning-based chromatic transformations can be optimized for downstream tasks, producing representations that better leverage RGB-pretrained networks \cite{Morra2021}. Importantly, a task-optimized chromatic representation may also provide a human-centric benefit: by increasing perceptual separability without occluding anatomy, it may reduce search effort and support more consistent interpretation in challenging cases.

In this work, we argue for a shift from purely ``automated diagnosis'' toward \emph{human-centric AI augmentation}. We propose \textbf{MammoColor}, an end-to-end framework that incorporates a \textbf{Task-Driven Chromatic Encoding (TDCE)} module trained jointly with a BI-RADS triage classifier. Unlike fixed pseudo-color colormaps or generic contrast enhancement, TDCE learns a risk-aware chromatic mapping that translates subtle pathological signatures (e.g., spiculated margins or distortion patterns) into visually salient chromatic cues, aiming to help radiologists ``visualize the invisible'' while preserving high-resolution anatomical context.

\noindent\textbf{Terminology.}
In this manuscript, we use \emph{chromatic encoding} as an umbrella term for mapping single-channel mammograms into a multi-channel (RGB) representation. \textbf{TDCE} denotes our learnable, task-driven \emph{encoding module} that generates \emph{TDCE-encoded} mammograms. \textbf{MammoColor} refers to the end-to-end architecture that couples the TDCE module with the downstream classifier and training strategy. The term \emph{pseudo-color} is reserved only for handcrafted, intensity-driven colormaps used as baselines.

\noindent\textbf{Key points}
\begin{itemize}
    \item We propose a clinically oriented \textbf{Task-Driven Chromatic Encoding (TDCE)} module within \textbf{MammoColor} to visually enhance dense-breast anatomy and occult mammographic findings.
    \item We demonstrate robustness through multicenter external validation across heterogeneous scanners and diverse patient populations, supporting generalizability under real-world domain shift.
    \item In a rigorous multi-reader, multi-case observer study, TDCE-encoded visualization improved diagnostic performance and increased inter-reader agreement, highlighting strong translational potential.
\end{itemize}

\section{Materials and Methods}\label{sec:2}

\begin{table*}[ht]
\centering
\small
\caption{Baseline characteristics of the study cohorts tailored to the hierarchical validation design.}
\label{tab:baseline}
\resizebox{\textwidth}{!}{%
\begin{tabular}{lcccccc}
\toprule
\multirow{2}{*}{\textbf{Characteristic}} & \textbf{Source Domain} & \multicolumn{2}{c}{\textbf{Domain Adaptation}} & \multicolumn{3}{c}{\textbf{Zero-Shot Generalization}} \\
\cmidrule(lr){2-2} \cmidrule(lr){3-4} \cmidrule(lr){5-7}
 & \textbf{VinDr} & \textbf{CBIS-DDSM} & \textbf{INBreast} & \textbf{Shenzhen} & \textbf{Liuzhou} & \textbf{Xuzhou} \\
\midrule
\textbf{Data Source} & FFDM & Scanned Film & FFDM & FFDM & FFDM & FFDM \\
\textbf{Patients (n)} & 5,000 & 1,566 & 115 & 598 & 101 & 466 \\
\textbf{Images (n)} & 19,995 & 3,032 & 410 & 1,170 & 185 & 921 \\
\textbf{Age (mean$\pm$SD)} & NR & NR & 57.4$\pm$13.5 & 49.0$\pm$10.2$^\dagger$ & 48.4$\pm$8.3$^*$ & 50.6$\pm$10.6 \\
\midrule
\multicolumn{7}{l}{\textbf{BI-RADS Category, n (\%)}} \\
\quad 0 & 0 (0.0) & 236 (7.8) & 0 (0.0) & 22 (1.9) & 0 (0.0) & 0 (0.0) \\
\quad 1 & 13,401 (67.0) & 3 (0.1) & 67 (16.3) & 0 (0.0) & 0 (0.0) & 0 (0.0) \\
\quad 2 & 4,676 (23.4) & 339 (11.2) & 220 (53.7) & 46 (3.9) & 8 (4.3) & 0 (0.0) \\
\quad 3 & 930 (4.7) & 405 (13.4) & 23 (5.6) & 30 (2.6) & 25 (13.5) & 6 (0.7) \\
\quad 4 & 762 (3.8) & 1,519 (50.1) & 43 (10.5) & 917 (78.4) & 140 (75.7) & 891 (96.7) \\
\quad 5 & 226 (1.1) & 530 (17.5) & 49 (12.0) & 155 (13.2) & 12 (6.5) & 24 (2.6) \\
\quad 6 & 0 (0.0) & 0 (0.0) & 8 (2.0) & 0 (0.0) & 0 (0.0) & 0 (0.0) \\
\midrule
\multicolumn{7}{l}{\textbf{Breast Density, n (\%)}} \\
\quad Fatty (A/B) & 2,008 (10.0) & 1,596 (52.6) & 282 (68.8) & NR & NR & NR \\
\quad Dense (C/D) & 17,987 (90.0) & 1,434 (47.3) & 127 (31.0) & NR & NR & NR \\
\midrule
\multicolumn{7}{l}{\textbf{Key Findings, n (\%)}} \\
\quad Mass & 1,014 (5.1) & 1,394 (46.0) & 108 (26.3) & 733 (62.6) & 47 (25.4) & 537 (58.3) \\
\quad Calcification & 366 (1.8) & 1,511 (49.8) & 308 (75.1) & 1,170 (100.0) & 174 (94.1) & 921 (100.0) \\
\quad Asymmetry$^\ddagger$ & 350 (1.8) & 52 (1.7) & 14 (3.4) & 79 (6.8) & 66 (35.7) & 427 (46.4) \\
\quad Distortion & 104 (0.5) & 150 (4.9) & 4 (1.0) & 131 (11.2) & 12 (6.5) & 146 (15.9) \\
\quad No Finding & 18,227 (91.2) & 0 (0.0) & 68 (16.6) & 0 (0.0) & 3 (1.6) & 0 (0.0) \\
\bottomrule
\multicolumn{7}{l}{\footnotesize NR: Not Reported. Percentages may not sum to 100\% due to multiple findings per image.} \\
\multicolumn{7}{l}{\footnotesize $^\dagger$ Age available for 597/598 patients. $^*$ Age available for 62/101 patients.} \\
\multicolumn{7}{l}{\footnotesize $^\ddagger$ Asymmetry category includes global, focal, and developing asymmetries.}
\end{tabular}%
}
\end{table*}

\subsection{Study Population and Datasets}\label{subsec:2.1}

A retrospective, multi-cohort study was performed to evaluate MammoColor, an end-to-end framework with a Task-Driven Chromatic Encoding (TDCE) module, for mammography-based screening triage in both algorithmic evaluation and human reading.  The study workflow comprised: (i) model development on a public dataset, (ii) cross-cohort validation on independent public and private datasets, and (iii) a multi-reader, multi-case (MRMC) observer study to assess clinical utility.

The development cohort was VinDr-Mammo\cite{Nguyen2023VinDrMammo} (full-field digital mammography [FFDM]; 5,000 patients) and was used for model training. To examine generalizability across imaging sources, two public datasets were used as adaptation cohorts: CBIS-DDSM\cite{Lee2017CBISDDSM} (scanned film mammography; 1,566 patients) and INBreast\cite{Moreira2012INbreast} (FFDM; 115 patients). For external validation, we assembled a multicenter private cohort from three independent institutions in China (Liuzhou, ShenZhen, and XuZhou), comprising 1,165 patients acquired between January 2015 and October 2024. Baseline characteristics and the role of each cohort in the study pipeline are summarized in Table~\ref{tab:baseline}.

The model was trained at the view level, treating each CC or MLO view as an independent sample. At inference, the breast-level score was defined as the maximum predicted probability across the paired CC and MLO views of the same breast. Cases were dichotomized into binary triage labels based on the final BI-RADS assessment category recorded in the radiology report. BI-RADS 1–3 (negative to probably benign) were labeled as non-suspicious (negative), whereas BI-RADS 4–5 (including subcategories 4A–4C; suspicious to highly suggestive of malignancy) and BI-RADS 6 (biopsy-proven malignancy) were labeled as suspicious (positive). Although BI-RADS 6 represents known cancers, these examinations were retained to assess whether the model preserves salient malignant morphology and visual cues. BI-RADS 0 examinations, which indicate an incomplete assessment requiring additional imaging, were excluded to ensure a definitive reference standard for the triage task.

Each mammographic view (craniocaudal [CC] or mediolateral oblique [MLO]) was treated as a separate sample during model training and evaluation. Views from the same breast shared the same breast-level label. Dataset splitting was performed at the patient level to avoid information leakage across partitions.

For prespecified subgroup analyses, breast density was categorized as non-dense (BI-RADS categories A and B) versus dense (BI-RADS categories C and D) in accordance with the American College of Radiology (ACR) BI-RADS Atlas \cite{ACR_BIRADS_v2025_Mammography}. Similarly, lesion types were stratified into four distinct categories—mass, calcification, architectural distortion, and asymmetry—following the standard morphological descriptors established by BI-RADS lexicon.

All mammograms were converted to 16-bit PNG format to preserve dynamic range. A standardized preprocessing pipeline was applied, including Otsu thresholding for background removal and cropping to the breast region of interest (ROI). Images were rescaled with the aspect ratio preserved, then padded to a uniform size, and intensity-normalized to [0,1] before network input.

\subsection{Task-Driven Chromatic Encoding Network}\label{subsec:2.2}

The TDCE module was embedded within the overall MammoColor network and trained end-to-end with the downstream BI-RADS triage classifier. Specifically, we implemented TDCE as a learnable chromatic encoding module that converts single-channel grayscale mammograms into three-channel RGB \emph{TDCE-encoded} images. Unlike handcrafted, intensity-driven colormaps, the TDCE module was optimized jointly with the classifier: the classification loss was backpropagated through TDCE so that the learned chromatic mapping was directly aligned with the screening triage endpoint (non-suspicious vs suspicious).

Architecturally, the TDCE module adopted a lightweight U--Net-style\cite{unet} encoder--decoder design. The encoder captured global context, while the decoder restored spatial details to preserve fine structures relevant to mammographic interpretation (e.g., microcalcifications and subtle architectural distortion). The resulting TDCE-encoded images were then fed into an ImageNet-pretrained ResNet-18\cite{resnet} backbone followed by a classification head to produce the probability of being suspicious.

During training, the ImageNet-pretrained ResNet-18 feature extractor in MammoColor was kept frozen. Each optimization step updated only the parameters of the TDCE module and the final MLP classification head, thereby isolating learning to the task-driven chromatic encoding while retaining the pretrained representation. After convergence, the learned TDCE module was fixed.

For external validation and clinical evaluation, the TDCE module was frozen to ensure a consistent and standardized output of TDCE-encoded images that preserves fine breast structural details.

For the grayscale baseline, we employed a standard transfer learning approach. The single-channel mammograms were replicated to three channels and input into a ResNet-18 backbone initialized with ImageNet weights.This configuration serves as a robust baseline as it represents the current state-of-the-art (SOTA) paradigm for adapting single-channel medical images to deep learning backbones pretrained on natural images. Crucially, unlike the TDCE branch where the backbone was frozen to test the encoding efficacy, the baseline ResNet-18 backbone was partially fine-tuned on the VinDr-Mammo dataset alongside the classification head to ensure a rigorous competitive standard. Performance of this grayscale baseline was then compared with MammoColor using TDCE-encoded inputs.

\begin{figure}[ht]
\centering
\includegraphics[width=1\textwidth]{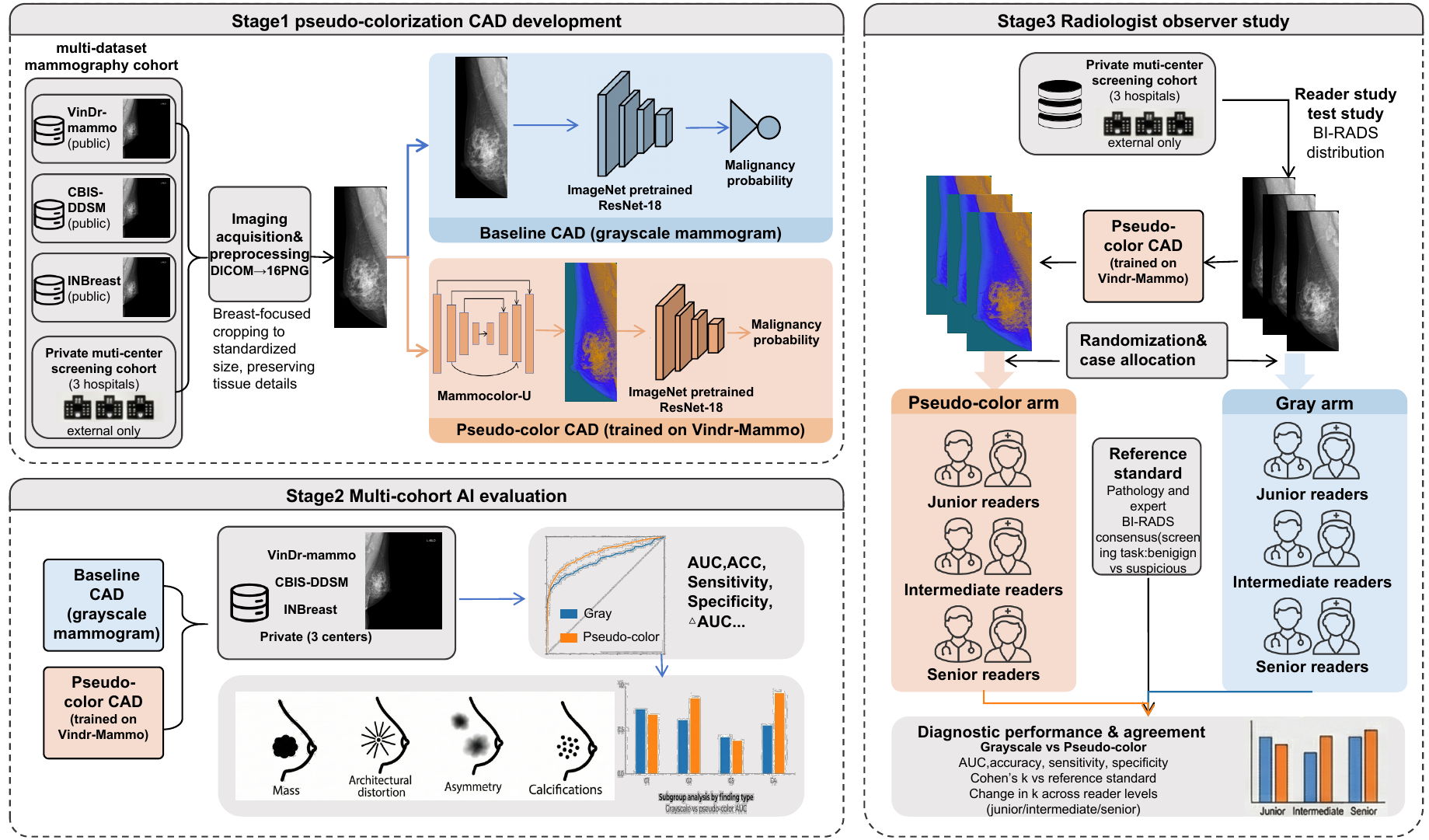}
\caption{Study overview. The workflow comprises three stages: (1) development of a baseline grayscale model and MammoColor (TDCE module + ImageNet-pretrained ResNet-18) after standardized preprocessing; (2) multi-cohort evaluation across public and private datasets; and (3) an external-only MRMC observer study with randomized case allocation to TDCE-encoded versus grayscale reading conditions, stratified by reader experience, using pathology and expert BI-RADS consensus as the reference standard.}
\label{fig:network}
\end{figure}

\subsection{Radiologist Observer Study Design}\label{subsec:2.3}
An MRMC observer study was conducted to evaluate the clinical utility of TDCE-encoded under a clinically realistic screening-triage workflow. Cases were sampled from the Shenzhen cohort, and an enriched set of 100 de-identified patients was assembled and balanced by the reference triage label extracted from the clinical report (50 non-suspicious, BI-RADS 1--3; 50 suspicious, BI-RADS 4--6). For each patient, the ipsilateral CC and MLO views were provided. All images were de-identified and underwent quality screening before inclusion.

Six radiologists were recruited and stratified by experience into three tiers: junior ($<5$ years), intermediate (5--10 years), and senior ($>10$ years), with two readers per tier. Each reader interpreted the same 100 cases under three reading conditions: standard grayscale mammograms, TDCE-encoded mammograms presented alone (TDCE-only), and a side-by-side display in which grayscale and TDCE-encoded images were shown concurrently with free switching between the two presentations. Readings were performed in three separate sessions (one condition per session), separated by a four-week washout period. To minimize period and order effects, the condition order was counterbalanced across readers using a three-condition Latin square (grayscale-only $\rightarrow$ TDCE-only $\rightarrow$ side-by-side; TDCE-only $\rightarrow$ side-by-side $\rightarrow$ grayscale-only; side-by-side $\rightarrow$ grayscale-only $\rightarrow$ TDCE-only), with readers assigned accordingly. Within each session, case order was independently randomized for each reader. Readers interpreted cases independently without access to clinical history or prior examinations; they were informed that the case set was enriched for suspicious findings but were not told the exact proportion.

For each case, readers provided both a binary triage assessment (non-suspicious vs. suspicious) and a BI-RADS category. For binary analysis, BI-RADS 1--3 were treated as non-suspicious and BI-RADS 4--6 as suspicious; BI-RADS 0 assessments, if present, were excluded. Primary endpoints were sensitivity, specificity, and accuracy under each reading condition, summarized overall and stratified by reader experience tier. Inter-reader agreement was quantified using Fleiss' kappa. Reading time was recorded for each condition; when a session was not completed in one uninterrupted sitting, timing was paused and resumed upon restart and aggregated to estimate total reading time per condition. Readers were aware that reading time was being recorded but were not provided with measured values.

\subsection{Statistical analysis}\label{subsec:2.4}

Model discrimination was assessed using receiver operating characteristic (ROC) analysis. The primary metric was the area under the ROC curve (AUC), reported with 95\% confidence intervals (CIs). For each cohort, AUCs for the grayscale baseline and TDCE models were compared using De Long's test.

To report clinically interpretable metrics, precision, sensitivity, and specificity were calculated using a fixed Youden decision threshold for all cohorts. Nonparametric bootstrap resampling was used at the \textbf{patient level} (2,000 resamples) to estimate 95\% CI for AUC and threshold-based metrics. For threshold-based comparisons between grayscale and TDCE evaluated on the same images, paired differences and corresponding 95\% CIs were derived from bootstrap replicates; McNemar's test was used as a sensitivity analysis for paired binary outcomes at the fixed threshold.

Performance was also evaluated in prespecified subgroups, including breast density (non-dense A/B vs dense C/D) and strata of lesions. Subgroup AUC were reported with 95\% CIs, and grayscale-versus-TDCE AUC differences within each subgroup were tested using De Long's test. Unless otherwise specified, subgroup analyses were considered exploratory.

The performance of the reader in the MRMC study was evaluated against the reference classification labels (BI-RADS 1--3 vs 4--6) of the clinical report. For each reading condition, sensitivity, specificity and accuracy were calculated per reader and summarized in general and by experience tier.

To account for the MRMC structure and repeated reads of the same cases across conditions, comparisons between reading conditions were performed using mixed-effects logistic regression, with reading condition as a fixed effect and random intercepts for reader and case. Separate models were fitted for: (i) overall correctness (correct vs incorrect classification) to compare accuracy; (ii) reference-positive cases only (correct positive vs miss) to compare sensitivity; and (iii) reference-negative cases only (correct negative vs false positive) to compare specificity.

All tests were two-sided, and $P<0.05$ was considered statistically significant. Analyses were performed using standard statistical software.

\begin{table}[t]
\small
\centering
\setlength{\tabcolsep}{3pt}
\renewcommand{\arraystretch}{1.05}
\caption{CAD performance with grayscale vs. TDCE-encoded inputs across datasets.
AUC is reported with 95\% confidence intervals in parentheses. $P$-values denote the statistical significance of the difference in AUC between Grayscale and TDCE-encoded inputs, with $P \leq 0.05$ considered statistically significant.}
\label{tab:cad_across_datasets}
\begin{tabular}{llcccccccc}
\hline
Dataset & Input & AUC (95\% CI) & Sens & Spec & BalACC & F1 & PPV & NPV & $p$-value \\
\hline

\multirow{2}{*}{VinDr}
  & Gray   & 0.7669 (0.7013--0.8232) & 0.5253 & 0.9453 & 0.7353 & 0.4078 & 0.3333 & 0.9745 & \multirow{2}{*}{$\leq 0.05$} \\
  & TDCE-encoded & 0.8461 (0.8032--0.8873) & 0.7273 & 0.8306 & 0.7789 & 0.2921 & 0.1827 & 0.9832 & \\[2pt]

\multirow{2}{*}{CBIS-DDSM}
  & Gray    & 0.7559 (0.7057--0.8060) & 0.8026 & 0.5668 & 0.6847 & 0.6630 & 0.5648 & 0.8039 & \multirow{2}{*}{0.4109} \\
  & TDCE-encoded & 0.7391 (0.6871--0.7873) & 0.7237 & 0.6544 & 0.6890 & 0.6528 & 0.5946 & 0.7717 & \\[2pt]

\multirow{2}{*}{INBreast}
  & Gray    & 0.8462 (0.6977--0.9560) & 1.0000 & 0.5769 & 0.7885 & 0.6452 & 0.4762 & 1.0000 & \multirow{2}{*}{0.3029} \\
  & TDCE-encoded & 0.9115 (0.8046--0.9753) & 1.0000 & 0.7308 & 0.8654 & 0.7407 & 0.5882 & 1.0000 & \\[2pt]

\multirow{2}{*}{Liuzhou}
  & Gray    & 0.5040 (0.3414--0.6647) & 0.8795 & 0.2778 & 0.5786 & 0.8639 & 0.8488 & 0.3333 & \multirow{2}{*}{$\leq 0.05$} \\
  & TDCE-encoded & 0.6329 (0.4832--0.7758) & 0.7952 & 0.5000 & 0.6476 & 0.8354 & 0.8800 & 0.3462 & \\[2pt]

\multirow{2}{*}{Shenzhen}
  & Gray   & 0.5379 (0.4548--0.6112) & 0.6357 & 0.4694 & 0.5525 & 0.7554 & 0.9307 & 0.1031 & \multirow{2}{*}{$\leq 0.05$} \\
  & TDCE-encoded & 0.6221 (0.5542--0.6845) & 0.5009 & 0.7959 & 0.6484 & 0.6595 & 0.9649 & 0.1246 & \\[2pt]

\multirow{2}{*}{Xuzhou}
  & Gray    & 0.6509 (0.5912--0.7106) & 0.8215 & 0.6667 & 0.7441 & 0.9009 & 0.9974 & 0.0235 & \multirow{2}{*}{0.132} \\
  & TDCE-encoded & 0.6785 (0.6135--0.7479) & 0.6280 & 1.0000 & 0.8140 & 0.7715 & 1.0000 & 0.0170 & \\
\hline
\end{tabular}
\end{table}

\section{Results}\label{sec:results}

\subsection{AI Performance Across Cohorts}\label{subsec:3.1}

To evaluate whether task-driven chromatic encoding (TDCE) improves mammography-based suspicion classification beyond conventional enhancement, we compared a grayscale baseline with the MammoColor across six cohorts spanning distinct imaging sources and clinical distributions (VinDr-Mammo, INBreast, CBIS-DDSM, and three independent Chinese multicenter cohorts). The primary endpoint was breast-level discrimination (AUC), where the breast-level score was defined as the maximum predicted probability across CC and MLO views. Secondary operating-point metrics at the Youden threshold (accuracy, sensitivity, specificity, PPV, NPV, F1, and balanced accuracy), together with 95\% confidence intervals and statistical tests, are summarized in Table \ref{tab:cad_across_datasets}.

On the development cohort (VinDr-Mammo), MammoColor significantly improved discrimination over the grayscale baseline (AUC 0.846 vs. 0.767; $\Delta$ AUC = +0.079; $P \leq 0.05$). At the Youden operating point, TDCE yielded higher sensitivity (0.727 vs. 0.525), whereas specificity and overall accuracy were modestly higher under grayscale.

Generalization performance varied across the two public domain-shift cohorts. On INBreast,  MammoColor showed a higher AUC than grayscale (0.912 vs. 0.846) with concordant trends in accuracy and balanced accuracy, although the difference was not statistically significant, consistent with the limited cohort size. In CBIS-DDSM, TDCE did not improve AUC relative to grayscale (Table \ref{tab:cad_across_datasets}). Notably, CBIS-DDSM comprises scanned-film mammograms, representing a larger domain shift from FFDM-based cohorts.

Across three independent external clinical cohorts from China (Liuzhou, Shenzhen, and Xuzhou), MammoColor achieved comparable or better discrimination than the grayscale baseline despite heterogeneity in acquisition settings and devices. The improvement was statistically significant in the Shenzhen cohort (AUC 0.622 vs. 0.538; $P< 0.05)$, while results for the remaining cohorts are reported in Table \ref{tab:cad_across_datasets}. Collectively, these multicohort findings support the robustness of TDCE across heterogeneous clinical settings.

\begin{figure}[t]
  \graphicspath{{figures/}}
  \centering
  \includegraphics[width=\linewidth]{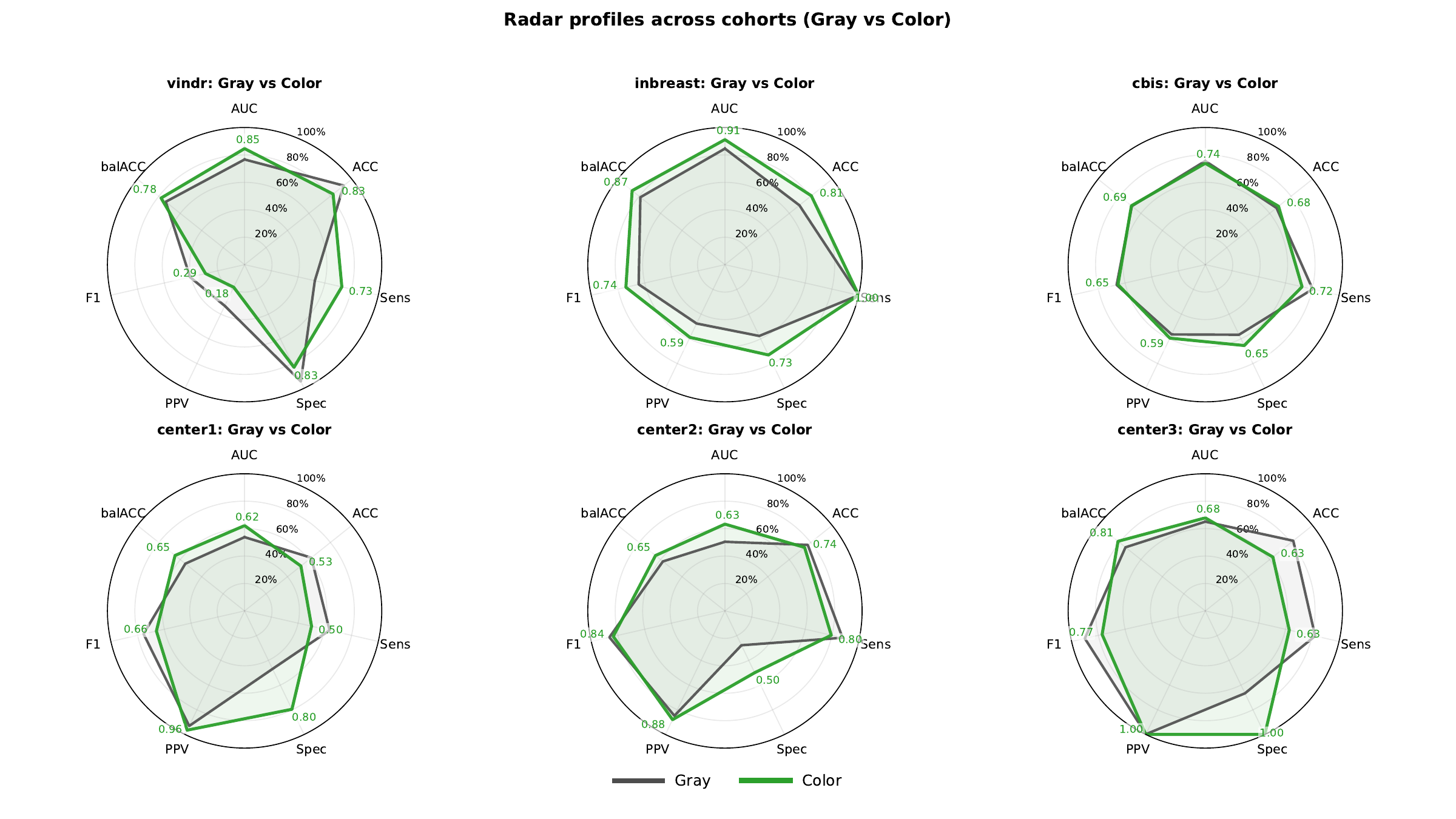}
  \caption{Radar profiles across cohorts (Gray vs TDCE-encoded).}
  \label{fig:radar_profiles}
\end{figure}

\subsection{Subgroup Analyses in VinDr-Mammo}\label{subsec:3.2}
Because density and lesion-type annotations were incomplete in parts of CBIS-DDSM and INBreast, prespecified subgroup analyses were performed primarily in VinDr-Mammo to ensure consistent labeling and interpretability. Subgroups included breast density (non-dense A/B vs. dense C/D) and four radiographic descriptors (calcification, mass, asymmetry, and architectural distortion).

In density-stratified analyses, MammoColor showed a significant advantage in dense breasts (C/D), increasing AUC from 0.749 to 0.835 ( $\Delta$ AUC = +0.086; $P \leq 0.05$). In non-dense breasts (A/B), AUC also increased (0.933 vs. 0.894), but statistical significance was not reached, consistent with the smaller number of positive cases in this subgroup.

In lesion-type stratification, MammoColor significantly improved performance for calcifications and masses. AUC increased from 0.875 to 0.948 for calcifications ($P \leq 0.05$) and from 0.743 to 0.838 for masses ($P \leq 0.05$). For asymmetry and architectural distortion, MammoColor showed consistent trends toward improved AUC, although differences did not reach statistical significance, likely reflecting limited positive counts. Full subgroup estimates are provided in Table\ref{tab:subgroup_vindr}.

\begin{table}[t]
\small
\centering
\setlength{\tabcolsep}{4pt}
\renewcommand{\arraystretch}{1.08}
\caption{Subgroup analyses in VinDr-Mammo. AUC is reported with 95\% confidence intervals in parentheses. Sensitivity and specificity are reported as point estimates. P-values (DeLong) indicate the statistical significance of the difference in AUC between Gray and TDCE-encoded within each subgroup.}
\label{tab:subgroup_vindr}
\begin{tabular}{lllc c c c}
\toprule
Subgroup & Pos/Neg & Modality & AUC (95\% CI) & Sens & Spec & AUC $p$ (DeLong) \\
\midrule

\multirow{2}{*}{Non-Dense}
  & \multirow{2}{*}{11/189}
  & Gray  & 0.894 (0.742--0.987) & 0.909 & 0.825 & \multirow{2}{*}{0.1296} \\
  &       & TDCE-encoded & 0.933 (0.847--0.991) & 0.909 & 0.921 & \\[2pt]

\multirow{2}{*}{Dense}
  & \multirow{2}{*}{88/1712}
  & Gray  & 0.749 (0.686--0.809) & 0.557 & 0.894 & \multirow{2}{*}{$\leq 0.05$} \\
  &       & TDCE-encoded & 0.835 (0.787--0.884) & 0.705 & 0.828 & \\[2pt]

\multirow{2}{*}{Calcification}
  & \multirow{2}{*}{47/1901}
  & Gray  & 0.875 (0.802--0.941) & 0.723 & 0.948 & \multirow{2}{*}{0.0375} \\
  &       & TDCE-encoded & 0.948 (0.914--0.979) & 0.830 & 0.935 & \\[2pt]

\multirow{2}{*}{Mass}
  & \multirow{2}{*}{59/1901}
  & Gray  & 0.743 (0.659--0.823) & 0.542 & 0.915 & \multirow{2}{*}{$\leq 0.05$} \\
  &       & TDCE-encoded & 0.838 (0.776--0.898) & 0.763 & 0.831 & \\[2pt]

\multirow{2}{*}{Asymmetry}
  & \multirow{2}{*}{19/1901}
  & Gray  & 0.775 (0.637--0.915) & 0.632 & 0.954 & \multirow{2}{*}{0.0755} \\
  &       & TDCE-encoded & 0.869 (0.759--0.946) & 0.737 & 0.861 & \\[2pt]

\multirow{2}{*}{Arch.\ Dist.}
  & \multirow{2}{*}{11/1901}
  & Gray  & 0.738 (0.539--0.926) & 0.636 & 0.846 & \multirow{2}{*}{0.3043} \\
  &       & TDCE-encoded & 0.796 (0.658--0.941) & 0.545 & 0.955 & \\

\bottomrule
\end{tabular}
\end{table}

\subsection{MRMC Observer Study}\label{subsec:3.3}
A multi-reader, multi-case (MRMC) observer study was conducted to examine whether TDCE-encoded may influence radiologists’ diagnostic performance. Overall accuracy was similar between the TDCE-encoded and grayscale conditions (0.82 vs. 0.81). Sensitivity was slightly lower under TDCE-encoded, whereas specificity showed a potential improving trend (0.96 vs. 0.90)that did not reach statistical significance ($P = 0.052$). Reader-stratified analyses suggested a non-significant tendency toward fewer false-positive calls under TDCE-encoded, particularly among junior readers, while senior readers maintained stable overall accuracy. Detailed MRMC endpoints and stratified results are reported in Table \ref{tab:reader_performance}.

\begin{table}[ht]
\centering
\caption{Diagnostic performance of radiologists (grayscale vs.\ TDCE-encoded) on the 48-case test subset.}
\label{tab:reader_performance}
\small
\setlength{\tabcolsep}{3pt} 

\begin{tabular}{lccc}
\toprule
\textbf{Reader / Condition} & \textbf{Accuracy} & \textbf{Sensitivity} & \textbf{Specificity} \\
\midrule
\multicolumn{4}{l}{\textit{Individual Readers}} \\
Junior, grayscale           & 0.81 & 0.63 & 1.00 \\
Intermediate, grayscale     & 0.77 & 0.67 & 0.88 \\
Senior, grayscale           & 0.85 & 0.88 & 0.83 \\
Junior, TDCE-encoded    & 0.79 & 0.63 & 0.96 \\
Senior, TDCE-encoded    & 0.85 & 0.75 & 0.96 \\
\midrule
\multicolumn{4}{l}{\textit{Aggregated Mean}} \\
Mean of grayscale           & 0.81 & 0.72 & 0.90 \\
Mean of TDCE-encoded    & 0.82 & 0.69 & 0.96 \\
\midrule
\textbf{P-value (Difference)} & \textbf{0.73} & \textbf{0.66} & \textbf{0.052} \\
\bottomrule
\multicolumn{4}{l}{\footnotesize Note: P-values assess the statistical significance of the difference between} \\
\multicolumn{4}{l}{\footnotesize grayscale and TDCE-encoded conditions.}
\end{tabular}
\end{table}

\subsection{Qualitative Analysis}\label{subsec:3.4} Q
To evaluate the perceptual utility of the TDCE-encoded images, we conducted a qualitative assessment involving representative cases of four major mammographic lesion types: masses, asymmetries, architectural distortions (AD), and calcifications. The expert review compared the diagnostic salience of TDCE-encoded visualizations against standard grayscale and inverted (negative-mode) mammograms.

\begin{figure}[tp]
  \centering
  \graphicspath{{figures/}}
  \includegraphics[width=\linewidth]{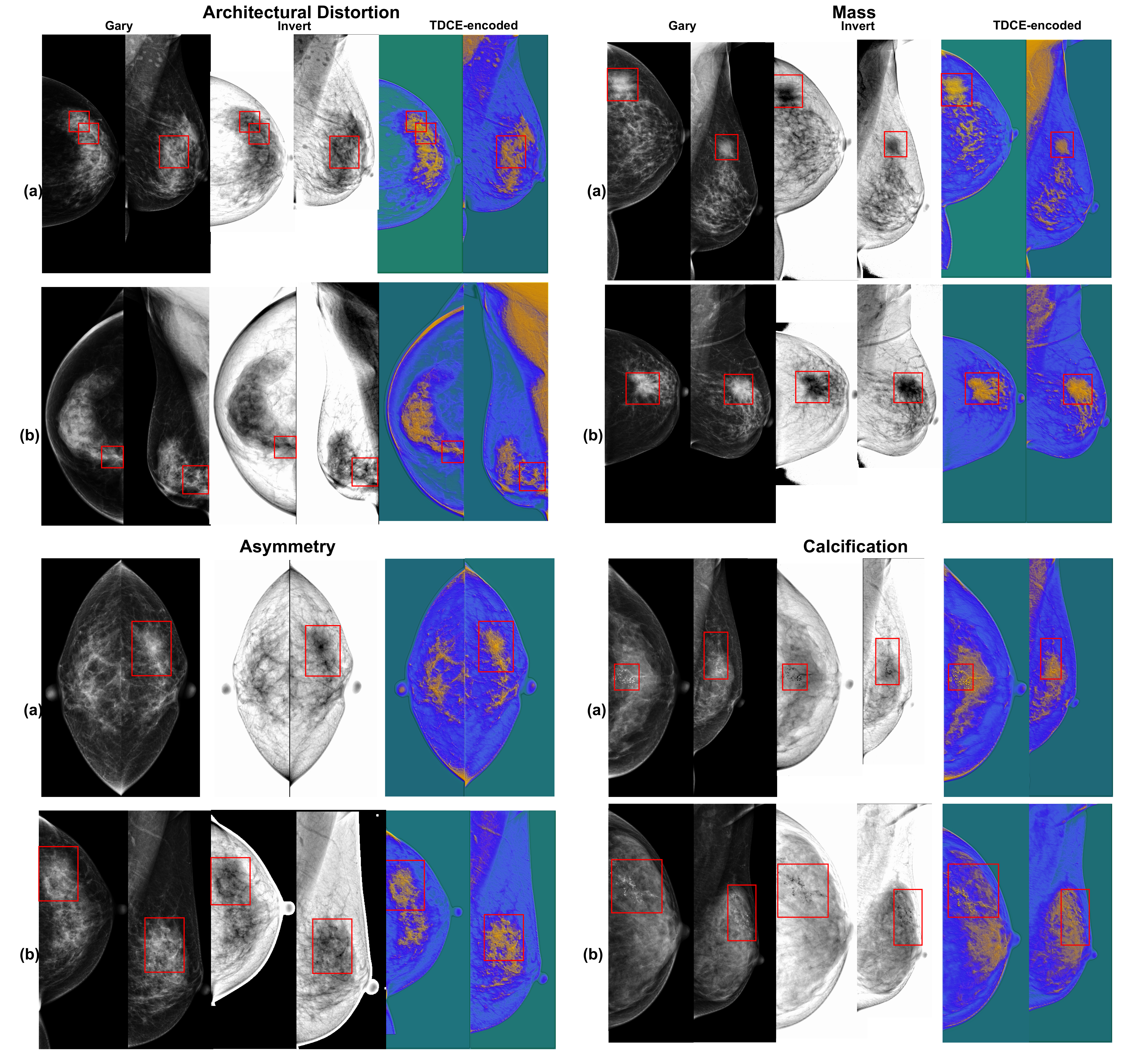} 
  \caption{Qualitative comparison of representative mammographic lesions across grayscale, negative-mode, and TDCE-encoded visualizations. The panels display four major lesion types (from left to right): masses, asymmetries, architectural distortions, and calcifications. For each type, Case (a) and Case (b) illustrate diverse pathological scenarios. The TDCE-encoded views demonstrate enhanced perceptual salience for masses and structural pulling, while grayscale and negative-mode remain complementary for fine calcification analysis.}
  \label{fig:qualitative_cases}
\end{figure}

\subsubsection{Masses and Asymmetries} 
For masses, TDCE significantly improved the visibility of morphological details. In cases of invasive carcinoma \ref{fig:qualitative_cases}mass (b), radiologists noted that while grayscale and negative-mode images clearly showed high-density masses with spiculed margins, TDCE provided superior contrast that highlighted micro-lobulations and the exact extent of the lesion boundaries. For asymmetries, especially in patients with high-grade ductal carcinoma in situ (DCIS), the task-driven chromatic mapping effectively quantized the differences in glandular density. The expert feedback indicated that asymmetries and associated multifocal masses were more "prominent and visually striking" in TDCE-encoded views compared to grayscale, facilitating the identification of subtle spiculated margins and mass contours that were previously obscured by overlapping parenchyma.

\subsubsection{Architectural Distortion} 
Architectural distortion often represents the most subtle sign of malignancy. In cases of invasive carcinoma \ref{fig:qualitative_cases}Architectural Distortion, grayscale and negative-mode visualizations showed radial pulling and focal convergence. However, TDCE-encoded images rendered the radial spicules with higher clarity, effectively isolating the lesion from the surrounding glandular background. The radiologists observed that TDCE provided a more definitive localization of the lesion center and a clearer demarcation of the associated irregular, high-density mass, whereas negative-mode display occasionally offered inferior visualization of these fine structural disruptions.

\subsubsection{Calcifications and Limitations}
While TDCE excels in highlighting structural and density-based abnormalities, its performance on fine calcifications varies. In cases of DCIS characterized by amorphous or fine pleomorphic calcifications \ref{fig:qualitative_cases}, grayscale and negative-mode images remained the preferred choice for assessing individual calcification morphology. The expert review suggested that while TDCE preserved the high-density characteristics of masses and dense clusters, it tended to blur the edges of individual microcalcifications or cause them to partially merge, making it difficult to differentiate between fine linear or pleomorphic subtypes. Conversely, standard grayscale and negative-mode displays were more effective at piercing through dense glandular tissue to show these granular details.

\subsubsection{Summary of Perceptual Impact}
Overall, the expert feedback confirms that TDCE-encoded visualization offers a "perceptual pop-out" effect for masses and architectural distortions, which are critical for screening triage. By augmenting the perceptual salience of density-based features and structural pulling, TDCE compensates for the inherent limitations of grayscale in dense breast phenotypes, although standard grayscale remain complementary for the fine-grained analysis of microcalcifications.

\section{Discussion}\label{sec:discussion}

The present study demonstrates that the MammoColor framework, incorporating a task-driven chromatic encoding (TDCE) module, can improve mammography-based screening triage across heterogeneous cohorts. Across six cohorts, MammoColor showed robust overall discrimination, with the most consistent gains observed in clinically challenging settings, including dense breasts and key lesion subtypes. Together with the observer-study findings, these results support the premise that the value of AI in mammography may extend beyond automated classification toward providing a complementary visual representation that helps surface subtle malignant cues that are otherwise masked by overlapping fibroglandular tissue.

A plausible explanation for the TDCE module’s advantage lies in representation-level feature enhancement enabled by end-to-end learning. In standard transfer-learning pipelines, grayscale mammograms are often replicated across RGB channels, leaving the network to infer semantic structure without additional representational degrees of freedom at the input level. In contrast, the TDCE module encourages the model to encode diagnostically relevant variations into a chromatic space, thereby increasing perceptual separability for both the network and the human reader. Rather than functioning as a post-hoc highlight, the resulting chromatic saliency can act as a perceptual “pop-out” cue that may reduce search effort and improve consistency, particularly in dense tissue where contrast is intrinsically limited.

Positioning relative to prior work. Unlike traditional image-enhancement techniques that rely on predefined, image-agnostic transformations\cite{colormap}, the MammoColor framework represents a shift toward a task-driven, learnable paradigm in which the visual representation is optimized end-to-end for the specific clinical objective of detecting malignancy. This coupling helps ensure that the resulting chromatic signals align with pathological risk rather than merely amplifying visual noise. Furthermore, while common domain generalization strategies attempt to bridge shifts across medical centers via feature-level alignment, MammoColor provides a more intuitive form of representation bridging at the earliest stage of the pipeline. By standardizing the diagnostic signal at the input level before deep feature extraction, the framework can complement existing methods and may offer a more stable foundation for cross-hospital deployment where acquisition varies substantially. Finally, unlike post-hoc heatmap\cite{Grad-CAM} explanations that may obscure critical clinical details, TDCE embeds diagnostic cues into the high-resolution image, enabling radiologists to inspect both the model’s cues and the underlying anatomy without visual obstruction.

The clinical relevance of TDCE lies in its potential to support screening workflows where the goal is to minimize missed cancers while managing the cognitive burden associated with false positives. In this context, TDCE is better viewed as an augmented interface than as a standalone “black-box” classifier. Our MRMC results suggested that TDCE-encoded reading tended to yield higher specificity without materially changing overall accuracy, consistent with the possibility that chromatic cues help readers reject false positives arising from glandular overlap or benign structures. At the same time, the algorithmic operating point can be tuned toward higher sensitivity (with modest specificity trade-offs), while the clinician provides contextual arbitration. This “triage-and-arbitration” framing emphasizes complementary strengths: algorithmic vigilance for subtle findings and clinician judgment for contextual filtering.

Several limitations warrant consideration. First, we observed attenuated or inconsistent gains in scanned-film cohorts (e.g., CBIS-DDSM), suggesting that TDCE may be less stable under larger domain shifts. This finding motivates systematic style calibration and/or targeted adaptation strategies to improve robustness across legacy imaging formats. Second, the MRMC study was performed on an enriched retrospective set; therefore, metrics such as PPV should not be directly extrapolated to population screening where prevalence is substantially lower. Finally, perceptual bias remains a potential concern: salient chromatic cues could inadvertently draw attention away from subtle findings outside highlighted regions, which calls for larger-scale human-factors evaluation and careful interface design.

Beyond the current scope of binary triage, task-driven chromatic encoding is scalable across the broader landscape of medical imaging. Future research will explore integrating TDCE into multi-task clinical workflows.

\section{Conclusion}\label{sec:conclusion}
In conclusion, this study introduces MammoColor, an end-to-end mammography framework incorporating a task-driven chromatic encoding (TDCE) module, to enhance diagnostic discrimination across heterogeneous clinical cohorts. By demonstrating consistent benefits in dense breasts and critical lesion subtypes, the proposed approach provides a practical interface for decision support that complements both automated triage and expert interpretation. The multicenter evaluation and observer-study findings support the clinical feasibility of this paradigm and motivate further prospective validation of TDCE-enabled, risk-aware medical image visualization.





\bibliography{main}

\end{document}